\documentclass{article}
\usepackage{ijcai25}

\usepackage{multirow} 
\usepackage{amssymb}
\usepackage[table,xcdraw]{xcolor}
\usepackage{times}
\usepackage{soul}
\usepackage{url}
\usepackage[hidelinks]{hyperref}
\usepackage[utf8]{inputenc}
\usepackage[small]{caption}
\usepackage{graphicx}

\usepackage{amsmath}
\usepackage{amsthm}
\usepackage{adjustbox}
\usepackage{booktabs}
\usepackage[linesnumbered, ruled]{algorithm2e}
\usepackage{appendix}

\usepackage[switch]{lineno}

\title{Contrastive Cross-Course Knowledge Tracing via \\ Concept Graph Guided Knowledge Transfer}

\author{
Wenkang Han\textsuperscript{\rm 1},
Wang Lin\textsuperscript{\rm 1},
Liya Hu\textsuperscript{\rm 1},
Zhenlong Dai\textsuperscript{\rm 1},
Yiyun Zhou\textsuperscript{\rm 1},
\\
Mengze Li\textsuperscript{\rm 1},
Zemin Liu\textsuperscript{\rm 1},
Chang Yao\textsuperscript{\rm 1},
Jingyuan Chen\textsuperscript{\rm 1,*}
\\
\textsuperscript{1}\small Zhejiang University \\
 wenkangh@zju.edu.cn, jingyuanchen@zju.edu.cn
}

\begin{document}

\maketitle
\renewcommand{\thefootnote}{}
\footnotetext{\textsuperscript{*}Corresponding authors.}


\begin{abstract}
Knowledge tracing (KT) aims to predict learners' future performance based on historical learning interactions. However, existing KT models predominantly focus on data from a single course, limiting their ability to capture a comprehensive understanding of learners' knowledge states. In this paper, we propose \textbf{TransKT}, a contrastive cross-course knowledge tracing method that leverages concept graph guided knowledge transfer to model the relationships between learning behaviors across different courses, thereby enhancing knowledge state estimation. Specifically, TransKT constructs a cross-course concept graph by leveraging zero-shot Large Language Model (LLM) prompts to establish implicit links between related concepts across different courses. This graph serves as the foundation for knowledge transfer, enabling the model to integrate and enhance the semantic features of learners' interactions across courses. Furthermore, TransKT includes an LLM-to-LM pipeline for incorporating summarized semantic features, which significantly improves the performance of Graph Convolutional Networks (GCNs) used for knowledge transfer. Additionally, TransKT employs a contrastive objective that aligns single-course and cross-course knowledge states, thereby refining the model's ability to provide a more robust and accurate representation of learners' overall knowledge states. Our code and datasets are available at \url{https://github.com/DQYZHWK/TransKT/}.
\end{abstract}

\section{Introduction}
\begin{figure}[t]
  \centering
  \includegraphics[width=\linewidth]{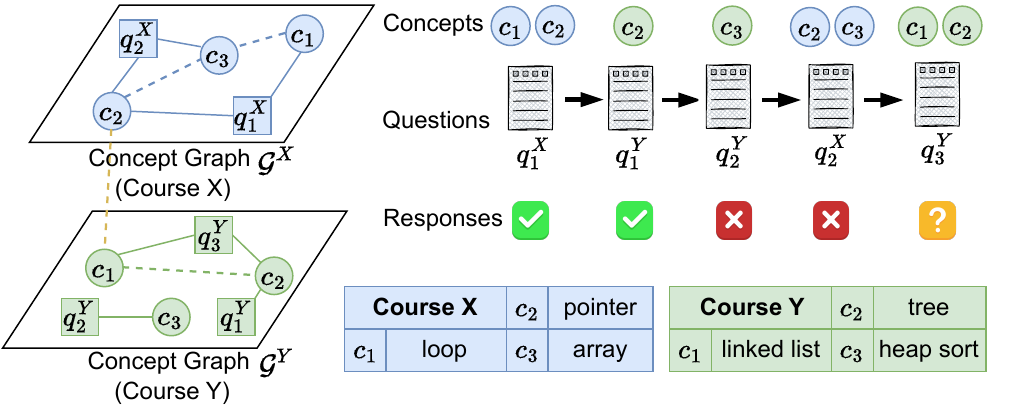}
  \caption{An illustrative example of Cross-Course Knowledge Tracing, where each question is associated with one or more concepts. $X$ denotes \textit{C Programming} course, and $Y$ denotes \textit{Data Structure and Algorithm Analysis} course.}
   \vspace{-0.4cm}
  \label{fig:toy_example}
\end{figure}

The popularity of massive open online courses (MOOCs) has expanded access to diverse educational resources, resulting in the accumulation of vast datasets on learner behaviors~\cite{ghosh2020context,wang2024eager,zhou2025revisiting}. This has led to the emergence of knowledge tracing (KT) as a critical technique for analyzing learners' behaviors, with the goal of providing personalized learning experiences.
KT aims to predict a learner's future performance (\textit{i.e.}, the probability of correctly answering new questions), by dynamically assessing their knowledge state based on historical interactions. 

Most existing KT models~\cite{ghosh2020context,huang2023towards,liu2023enhancing,zhou2024cuffkt,zhou2025disentangled} are developed based on data from a single course, limiting their ability to capture overall knowledge state across different courses. This paradigm often results in a fragmented understanding, as in practice, learners frequently engage with multiple courses simultaneously~\cite{simamora2020challenges,hu2023ptadisc}. The cumulative effect of these varied learning experiences plays a crucial role in shaping a comprehensive knowledge state. In such cases, a single-course focus may lead to unstable and suboptimal predictions, overlooking the interdependencies and transferability of knowledge across different courses. For example, as shown in Figure~\ref{fig:toy_example}, while Course $X$ and Course $Y$ might not share direct concepts, implicit connections (indicated by the yellow dashed line) between their respective concepts could significantly influence a learner's overall proficiency. 
This interconnectedness highlights the importance of considering learners' experiences across multiple courses to gain a comprehensive understanding of their knowledge states.

To bridge this gap, we introduce a novel task called cross-course knowledge tracing (CCKT), which aims to enhance knowledge state estimation by leveraging the relationships between learning behaviors across different courses. For example, by considering a learner's historical interactions in both Course $X$ and Course $Y$, the goal of CCKT is to predict the probability that the learner can correctly answer the future question, such as $q_3^Y$.

However, CCKT is not a trivial task and presents significant challenges. Firstly, the connections between concepts are often sparse, especially when comparing concepts from different courses. This scarcity makes it difficult to transfer a learner’s proficiency in one concept to potentially related concepts in another course. Secondly, while integrating knowledge states across multiple courses provides a more comprehensive view, it can also introduce noise from unrelated interactions. Therefore, effectively synthesizing learners' knowledge states within and across courses to achieve a robust representation remains another key challenge in CCKT.

\begin{figure*}[t]
\centering
\includegraphics[width=1\textwidth]{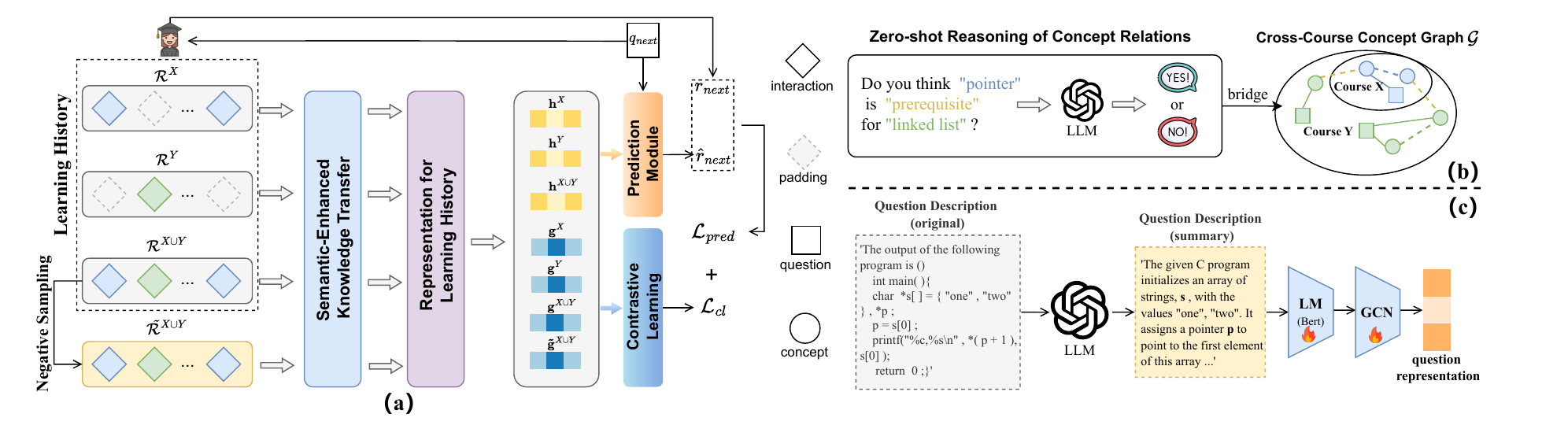}
\vspace{-1.5em}
\caption{The overview of the proposed TransKT model is shown in (a). The construction of the cross-course concept graph is illustrated in (b). The pipeline of semantic-enhanced knowledge transfer is depicted in (c).}
\vspace{-0.5em}
\label{fig:mainwork}
\end{figure*}

To address these challenges, we propose a contrastive cross-course knowledge tracing model via concept graph guided knowledge transfer (denoted as \textbf{TransKT}). As shown in Figure~\ref{fig:mainwork}, TransKT predicts learners' performance on future questions in a course by analyzing their historical learning interactions across multiple courses. Specifically, to address the first challenge, we introduce a cross-course concept graph construction module that establishes implicit links between intra-course and inter-course concepts by an advanced Large Language Model (LLM). Furthermore, based on the constructed cross-course concept graph, we propose a semantic-enhanced knowledge transfer module, which leverages an LLM-to-LM\protect\footnotemark pipeline to summarize and extract rich semantic features, which are then utilized to enhance the performance of Graph Convolutional Networks (GCNs) in cross-course knowledge transfer. 
Additionally, we derive the learner's knowledge state based on their learning history. Finally, to overcome the second challenge, we propose a cross-course contrastive objective with hybrid hard negative sampling strategy to maximize the mutual information between single-course and cross-course knowledge states. This objective aims to encourage the correlation of learners' knowledge states within and across courses for a more robust knowledge state representation. To demonstrate the validity of our model, we utilize the publicly available PTADisc~\cite{hu2023ptadisc} dataset to derive three cross-course KT datasets and use the state-of-the-art KT models for comparison.

The contributions of this paper can be summarized as:
\begin{itemize}
    \item \textbf{Motivation.} This study advances cross-course knowledge tracing, underscoring the critical importance of modeling knowledge acquisition across multiple courses. To the best of our knowledge, the proposed TransKT model represents the first systematic effort to investigate knowledge tracing within a multi-course context, offering novel insights and methodologies for this unexplored yet significant area.
    \item\textbf{Method.} We introduce TransKT, a novel model for cross-course knowledge tracing designed to address the challenges of sparse connections and potential noise in cross-course integration. Firstly, we construct a cross-course concept graph with aligned semantics using an LLM-to-LM pipeline, enabling semantic-based knowledge transfer across courses via graph convolutional networks (GCNs). Secondly, we propose a cross-course contrastive objective that aligns single-course knowledge state representations with their cross-course counterparts, enhancing the robustness and coherence of the synthesized knowledge states. Additionally, as a content-based approach, TransKT allows new questions or concepts to be incorporated without any additional training, offering a significant advantage over traditional ID-based knowledge tracing methods.
    \item \textbf{Experiments.} Extensive experiments on three cross-course knowledge tracing datasets, demonstrating the superiority of TransKT over state-of-the-art baselines.
\end{itemize}

\section{Related works}
\subsection{Deep Learning Based Knowledge Tracing}
Inspired by deep learning~\cite{lecun2015deep}, recent knowledge tracing models generally apply deep learning technology.
DKT~\cite{piech2015deep} stands out as a representative method, employing Long Short-Term Memory to predict the probability of correct responses at each time step.
Inspired by memory-augmented neural networks, DKVMN~\cite{zhang2017dynamic} directly predicts learners' knowledge mastery levels based on the values of a dynamic memory matrix, constituting an extension method within the DKT framework. Furthermore, several studies~\cite{ghosh2020context,chen2023improving} have sought to integrate attention mechanisms into KT models following the emergence of the transformer architecture~\cite{vaswani2017attention}. The basic idea of these methods is to assign different attention weights to questions in a series of interactions. Additionally, recent studies~\cite{lee2022contrastive,yin2023tracing,zhou2025disentangled} have considered incorporating contrastive learning into the training of KT models to maintain the stability of knowledge state diagnostics. However, existing KT methods only consider learners' interactions within a single course, overlooking the transfer of knowledge between courses. 

\subsection{Text-attributed Graph Representation Learning}
Text-attributed graph (TAG) representation learning is an emerging field that integrates graph based learning with natural language processing to effectively utilize text attributes in graph-structured data. Recent research has used deep embedding techniques, leveraging pre-trained language models (LMs) like BERT~\cite{devlin2018bert} to generate rich node embeddings that encapsulate the semantic depth of text attributes. Approaches such as TextGNN~\cite{zhu2021textgnn} and GIANT~\cite{chien2021node} have demonstrated significant performance improvements by integrating LM-based embeddings with graph neural networks (GNNs). The incorporation of large language models (LLMs) such as ChatGPT~\cite{mann2020language}, presents new opportunities for enhancing TAG-related tasks~\cite{lv2024semantic,dai2024mpcoder,wu2024semantic,guo2025efficient}. TAPE~\cite{he2023harnessing} is a representative method that prompts an LLM for zero-shot classification explanations and then using an LLM-to-LM interpreter to translate these explanations into informative features for enhancing GNN performance. Although these strategies perform well in TAG-related tasks, the application of this idea to cross-course knowledge tracing, where the concept graphs of different courses are independent and the question textual contents of different courses vary greatly in form, has not been explored.
\section{Preliminary}
\label{problem statement}
\paragraph{Knowledge Tracing (KT)} 
involves tracing learners' knowledge states and predicting their future performance over time. Formally, assuming there is a course with a total of $n$ learners, $m$ questions, and $k$ concepts, which can be denoted as $\mathcal{S}=\left\{ s_1,s_2,\ldots,s_n \right\}$, $\mathcal{Q}=\left\{ q_1,q_2,\ldots,q_m \right\} $, and $\mathcal{C}=\left\{ c_1,c_2,\ldots,c_k \right\} $ respectively. 
The interaction records of the learner are denoted as $\mathcal{R}$, which is a sequence of tuples $\left( q,r \right)$, where $q\in\mathcal{Q}$ and $r$ indicates the binary correctness of the learner's response to the question $q$. 
Each question is associated with one or more concepts. To simplify the explanation, we use $\{c\}$ to represent the set of concepts associated with question $q$. KT aims to predict the probability that a learner will correctly answer the next question $q \in \mathcal{Q}$ by utilizing historical interaction records $\mathcal{R}$.

\paragraph{Cross-Course Knowledge Tracing (CCKT)} focuses on predicting a learner's performance on a new question based on their interleaved interaction records from \textit{multiple courses}. 
In a simplified scenario where learners' interaction records involve two courses, denoted as $X$ and $Y$, given ${\mathcal{R}}^X=[(q_1^X,r_1^X),\cdots, (q_{n_X}^X,r_{n_X}^X)]$ and ${\mathcal{R}}^Y=[(q_1^Y,r_1^Y),\cdots,(q_{n_Y}^Y,r_{n_Y}^Y)]$, representing the interaction records of the learner in courses $X$ and $Y$, respectively. 
The interaction records from both courses are merged in chronological order, forming $\mathcal{R}^{X \cup Y}=[(q_1^X,r_1^X),\cdots,(q_1^Y,r_1^Y),\cdots,(q_{n_Y}^Y,r_{n_Y}^Y)]_{n_X+n_Y}$.
We use a padding strategy\footnote{An example of padding for an interaction sequence of length 5:\\ $\mathcal{R}^X=[( q_{1}^{X},r_{1}^{X}),\text{pad},\text{pad},(q_{2}^{X},r_{2}^{X}),(q_{3}^{X},r_{3}^{X} )]$; \\$\mathcal{R}^Y=\text{[}\text{pad},(q_{1}^{Y},r_{1}^{Y}),(q_{2}^{Y},r_{2}^{Y}),\text{pad},\text{pad}\text{]}$; \\$\mathcal{R}^{X\cup Y}=\text{[}(q_{1}^{X},r_{1}^{X}),(q_{1}^{Y},r_{1}^{Y}) ,(q_{2}^{Y},r_{2}^{Y}),(q_{2}^{X},r_{2}^{X}),(q_{3}^{X},r_{3}^{X}) \text{]}$}
to align the interaction records of $\{\mathcal{R}^X,\mathcal{R}^Y,\mathcal{R}^{X \cup Y}\}$.
CCKT aims to predict the probability that the learner will correctly answer the next question $q_{n_X+1}^X$ (or $q_{n_Y+1}^Y$) based on the observed interaction records, which can be formulated as $p(r_{n_X+1}^X\text{=1} |\mathcal{R}^X,\mathcal{R}^Y,\mathcal{R}^{X \cup Y}, q_{n_X+1}^X )$ (or $p(r_{n_Y+1}^Y\text{=1} |\mathcal{R}^X,\mathcal{R}^Y,\mathcal{R}^{X \cup Y}, q_{n_Y+1}^Y )$).

\section{Methodology}
In this section, we present the overview of our TransKT model (Shown in Figure~\ref{fig:mainwork}(a)). Initially, a cross-course concept graph is constructed by predicting concept relations using zero-shot prompting Large Language Models (LLMs) (§\ref{sec:3-1}). Subsequently, based on the constructed graph, TransKT incorporates a semantic-enhanced knowledge transfer module (§\ref{sec:3-2}) that utilizes LLMs to extract semantic information as features, which are then utilized to enhance the performance of Graph Convolutional Networks (GCNs) in facilitating cross-course knowledge transfer. Furthermore, TransKT derives the learner's knowledge state based on their learning history(§\ref{sec:3-3}). Finally, TransKT incorporates a cross-course contrastive objective (§\ref{sec:3-4}) which maximizes mutual information between single-course and cross-course knowledge states to learn more robust representations for prediction. The prediction and training process of the framework are introduced in §\ref{sec:3-5}.

\subsection{Cross-Course Concept Graph Construction}
\label{sec:3-1}
In modern education, concept graphs (CGs) are widely used as powerful tools for organizing information, offering learners a more intuitive understanding of links between concepts~\cite{ain2023automatic}. By leveraging these links, knowledge tracing can uncover connections among learning records~\cite{yang2021gikt,liu2020improving,yu2024rigl}. However, in cross-course scenarios, the CGs of different courses are independent of each other. This independence hinders the transfer of knowledge between courses, making it challenging for cross-course knowledge tracing model to identify valuable learning records from other courses. To address this issue, we propose a zero-shot link prediction method to construct a cross-course concept graph $\mathcal{G}$.

As shown in Figure~\ref{fig:mainwork}(b), the cross-course concept graph, includes two types of nodes: questions and concepts, and two types of links: explicit question-concept links (solid line) and implicit concept-concept links (dotted line). As defined in the Preliminary (§\ref{problem statement}), the explicit question-concept links are predefined (\textit{i.e.}, the links between $q$ and $\{c\}$). To derive the implicit concept-concept links, we draw inspiration from previous work~\cite{yang2024graphusion} and identify four types of candidate relations ($\textit{i.e.}$, ``Prerequisite\_of'', ``Used\_for'', ``Hyponym\_of'', ``Part\_of''). 
We then design a prompt, $pro_r$, to leverage the robust zero-shot reasoning capabilities of large language models (LLMs)~\cite{mann2020language} for determining pairwise relations between concepts. The core components of $pro_r$ include the course names, the definition and description of the dependency relation to be predicted, and the query concepts. If a pair of concepts satisfies one of the relation types, an implicit concept-concept link is established in the cross-course concept graph $\mathcal{G}$. 

\subsection{Semantic-Enhanced Knowledge Transfer}
\label{sec:3-2}
Previous work~\cite{tong2022introducing} endeavors to integrate semantic information into knowledge tracing. However, in real-world scenarios, the textual content of questions and concepts can often be overly specific or overly abstract, making it challenging to extract meaningful semantic information. Therefore, we propose a semantic-enhanced approach to knowledge transfer in this section. \textbf{This approach is applicable not only to multi-course settings but also demonstrates effectiveness in single-course contexts, particularly when there are variations in the content structure across different questions.} The approach includes: 1) explanation generation with LLMs, 2) semantic feature encoding, and 3) semantic knowledge propagation.
\subsubsection{Explanation Generation with LLMs}
In educational settings, each question and concept is accompanied by specific textual content, including question descriptions and concept names. The diverse forms of this textual content make it challenging to directly extract effective semantic features with language models. For instance, in computer science course, question descriptions often contain intricate code, while concept names are typically concise and abstract. To align these different types of textual content, as shown in Figure~\ref{fig:mainwork}(c), TransKT employs an ``open-ended'' method to query large language models (LLMs)~\cite{mann2020language}, leveraging their general knowledge and powerful reasoning capabilities to summarize diverse textual content, which can be formulated as:
\begin{equation}
x_{sum} =
\begin{cases} 
\text{LLM}(pro_q:x_{ori}) & \text{if } x \in \mathcal{Q} \\
\text{LLM}(pro_c:x_{ori}) & \text{if } x \in \mathcal{C}
\end{cases},
\end{equation}
where $x_{ori}$ denotes the original textual content, $x_{sum}$ denotes the summarized textual content, and $\text{LLM}(\cdot)$ denotes the \text{LLM} interface. Prompts $pro_q$ and $pro_c$ are tailored respectively for question descriptions and concept names.

\subsubsection{Semantic Feature Encoding}
After obtaining $x_{sum}$, the next step is to convert these text-based outputs into fixed-length semantic features. This transformation facilitates the discovery of semantic relationships within educational content, particularly for questions and concepts across different courses. 
To achieve this, we fine-tune a smaller language model (LM) to serve as an $\textit{interpreter}$ for the outputs of the LLM. This process extracts the most valuable and relevant semantic features from $x_{sum}$ for knowledge tracing, which can be formulated as:
\begin{equation}
    \mathbf{x}=\text{LM}_{\theta}(x_{sum}) \in \mathbb{R}^{D},
\end{equation}
where $\text{LM}(\cdot)$ denotes a language model based on the transformer structure, such as RoBERTa~\cite{liu2019roberta}. Here, $\theta$ denotes the parameters of the LM, and $\mathbf{x}$ denotes the semantic features of a question or concept.

\subsubsection{Semantic Knowledge Propagation}
\label{sec:knowledge propagation}
Following the construction of the cross-course concept graph, $\mathcal{G}$, and the extraction of semantic features $\mathbf{x}$ for nodes within $\mathcal{G}$, knowledge transfer across courses is achieved through the utilization of graph convolutional networks (GCNs).
Specifically, we utilize GraphSAGE~\cite{hamilton2017inductive} by stacking multiple GCN layers to encode higher-order neighborhood information. At each layer, the representation of each node is updated by considering both its own semantic feature and those of its neighboring nodes. We denote the feature of node $i$ in the graph as $\mathbf{x}_i$, and the set of its neighbor nodes as $\mathcal{N}_i$. The $l$-th GCN layer can be expressed as:
\begin{equation}
    \mathbf{x}_i^l=\mathrm{ReLU}(\frac{1}{|\mathcal{N}_i|}\sum_{j \in \mathcal{N}_i \cup \{i\}} w^l\mathbf{x}_j^{l-1}+b^l).
\end{equation}

After knowledge propagation by GCN, we get the enhanced feature of questions and concepts, denoted as $\mathbf{q}$ and $\mathbf{c}$ respectively. We then combine $\mathbf{q}$ with the representation of the response to indicate interaction as:
\begin{equation}
\mathbf{q}^{r}=\mathbf{q}+\mathbf{r},
\end{equation}
where $\mathbf{r}=x_r\cdot \mathbf{W}_r$, with $x_r$ being a 2-dimensional one-hot vector indicating the correctness of the answer, and $\mathbf{W}_r \in \mathbb{R}^{2 \times D}$ representing a learnable weight parameter.
\label{sec:emb}

\subsection{Representation for Learning History}
\label{sec:3-3}
The process of knowledge acquisition is inherently heterogeneous due to the diverse characteristics of learners. In the context of knowledge tracing, a learner's historical learning data can serve as a reflection of these distinct traits in the acquisition of knowledge. So, we adopt the attention function to obtain context-aware interaction representations from the learner's historical learning records. The interaction level knowledge state $\mathbf{h}_{t+1}$, achieved after the learner completes the $t$-th learning interaction, is denoted as:
\begin{align*}
\mathbf{h}_{t+1}=\mathrm{SelfAttention}(Q,K,V),  \\
    Q={\mathbf{q}_{t+1}},K=\left\{ \mathbf{q}_1,\ldots,\mathbf{q}_t\right\},V=\left\{ \mathbf{q}^{r}_1,\ldots,\mathbf{q}^{r}_t\right\}.
\end{align*}
Then, an average pooling layer $\text{pool}(.)$ is used to represent the entire interaction history, which can be formatted as:
\begin{align}
    \mathbf{g}=\text{pool}(\mathbf{h}_{1:T}).
\end{align}
Here, $T$ represents the length of the learning history. The final output, $\mathbf{g} \in \mathbb{R}^{D}$ is utilized in the subsequent contrastive learning process.

\subsection{Cross-Course Contrastive Objective}
\label{sec:3-4}
As mentioned earlier, the single-course knowledge state focuses solely on the learning behaviors related to a particular course, potentially resulting in unstable and suboptimal predictions. Additionally, the cross-course knowledge state offers insights into multiple courses, but may also introduce noise from interactions with other courses. Therefore, it is essential to jointly learn both the learner's single-course and cross-course knowledge state in order to make more accurate predictions. Drawing on the concept of mutual information maximization~\cite{becker1996mutual,hjelm2018learning}, we propose a cross-course contrastive objective to push together the single-course and cross-course knowledge state of the same learner. Additionally, inspired by contrastive learning based KT methods~\cite{lee2022contrastive,yin2023tracing}, we propose a hybrid hard negative sampling strategy designed specifically for cross-course scenarios to further enhance the discriminative capability of the contrastive learning process.

\subsubsection{Hybrid Hard Negative Sampling}
\label{sec:negative}
The hybrid hard negative sampling module contains two strategies, namely the \textit{response flip strategy} and the \textit{interaction replace strategy}.

For the response flip strategy, we randomly flip learners' responses, which can be formatted as: 
\begin{equation}
    \mathcal{\tilde{R}}=\{ (q_i,\tilde{r}_i)\},
\end{equation}
where $i$ indicates the index of randomly selected interactions.

For the interaction replace strategy, for each correct interaction, we replace the question with an easier one and mark the response as incorrect, and vice versa. The hard sample generated by this strategy can be formatted as:
\begin{equation}
    \mathcal{\tilde{R}}=\{ (\mathcal{F}(q_i),\tilde{r}_i)\},
\end{equation}
where $i$ indicates the index of randomly selected interactions and $\mathcal{F}(\cdot)$ denotes the application of corresponding replacement operations. 

\subsubsection{Cross-Course Contrastive Learning}
\label{sec:infomax}
Motivated by maximizing mutual information, we propose a cross-course contrastive objective to push KT model to capture the shared and distinct information across courses. This objective aim to maximize the mutual information between the local (single-course) and global (cross-course) features of the same learner. Taking course $X$ as an example, we seek the single-course knowledge state $\mathbf{g}^X$ to be relevant to the cross-course knowledge state $\mathbf{g}^{X \cup Y}$ but irrelevant to the negative representation $\tilde{\mathbf{g}}^{X \cup Y}$. Therefore, the cross-course contrastive objective for course $X$ is formulated as:
\begin{equation}
\label{MIloss}
    \mathcal{L}_{cl}^X={-}( \log \mathcal{D}^X( \mathbf{g}^{X},\mathbf{g}^{X \cup Y} ) +\log ( 1-\mathcal{D}^X( \mathbf{g}^X,\mathbf{\tilde{g}}^{X \cup Y})  ) ),
\end{equation}
where $\mathcal{D}^X$ is a binary discriminator that scores local and global representation pairs through a bilinear mapping function:
\begin{equation}
    \mathcal{D}^X( \mathbf{g}^{X},\mathbf{g}^{X \cup Y}) =\mathrm{Sigmoid} {(  \mathbf{g^{X}} \mathbf{W}_{MI}^{X} ( \mathbf{g^{X \cup Y}} )^T ) }.
\end{equation}
Here, $\mathbf{W}_{MI}^{X} \in \mathbf{R}^{D \times D}$ is a learnable weight matrix. As mentioned in previous work~\cite{cao24contrastive,li2023contrastive}, the binary cross-entropy loss in Equation~(\ref{MIloss}) serves as an effective mutual information estimator.


\begin{table}[t]
\centering
\begin{adjustbox}{max width=0.85\columnwidth}
\begin{tabular}{lrrr}
\toprule
\textbf{Dataset} & \textbf{Java\&Python} & \textbf{C\&DS}  & \textbf{CS\&MA} \\
\midrule
\textbf{\#Records} & 1,129,999 & 1,800,066  & 282,135 \\
\textbf{\#Learners} & 7,770 & 12,275 & 2,431 \\
\textbf{\#Questions} & 5,734/7,562 & 11,934/7,624 & 5,870/1,386\\
\textbf{\#Concepts} & 360/364 & 362/323 & 359/140 \\
\bottomrule
\end{tabular}
\end{adjustbox}
\caption{CCKT dataset statistics.}
\label{tab:dataset_info}
\end{table}

\subsection{Prediction and Training}
\label{sec:3-5}

\subsubsection{Cross-Course Prediction Objective} 
In cross-course scenarios, when predicting a learner's performance on a new question, we simultaneously utilize the knowledge state from both cross-course and single-course perspectives to form the joint knowledge state. Taking course $X$ as an example, the joint interaction level knowledge state can be formatted as:
\begin{equation}
\label{eq:CCState}
    \hat{\mathbf{h}}_{t+1}^X = \eta\cdot \mathbf{h}^{X \cup Y}_{t+1}+ ( 1-\eta )\cdot\mathbf{h}_{t+1}^{X},
\end{equation}
where $\eta \in [0,1]$ is a hyperparameter to balance the influence of the cross-course $\mathbf{h}^{X \cup Y}_{t+1}$ and single-course knowledge state $\mathbf{h}_{t+1}^{X}$ at timestep $t+1$. This joint knowledge state $\hat{\mathbf{h}}_{t+1}^X$ is then used to predict the learner's performance on a new question in course $X$.
\label{sec:predicate}
Specifically, we concatenate $\hat{\mathbf{h}}_{t+1}^X$ with the question representation $\mathbf{q}_{t+1}$ and use the binary cross-entropy loss. The prediction output $\hat{r}_{t+1}^X$ and the loss function $\mathcal{L}_{pred}^{X}$ are calculated as:

\begin{equation}
    \hat{r}_{t+1}^X=\mathrm{ReLU} ( \mathbf{W}^{X}\cdot (\hat{\mathbf{h}}_{t+1}^X \oplus \mathbf{q}_{t+1})  ),
    \label{eq:predicate}
\end{equation}
\begin{equation}
    \mathcal{L}_{pred}^{X}=-\sum_{t}{( r_{t}^X\log \hat{r}_{t}^X+ (1-r_{t}^X)\log (1-\hat{r}_{t}^X))},
\end{equation}
where $\mathbf{W}^X \in  \mathbb{R}^{D \times 2D}$ represents the learnable weight parameter.

\subsubsection{Model Training}
During training of TransKT, a regularization factor $\lambda$ is utilized to balance the prediction loss and contrastive learning loss. The final loss function is:
\begin{equation}
    \mathcal{L}=\lambda ( \mathcal{L}_{pred}^{X}+\mathcal{L}_{pred}^{Y} ) +( 1-\lambda ) ( \mathcal{L}_{cl}^X +\mathcal{L}_{cl}^Y ). 
\end{equation}


\section{Experiments}

\begin{table*}[t]
\centering
\vspace{-1em}
\begin{adjustbox}{max width=\textwidth}
\begin{tabular}{lcccccccccccc}
\toprule
\multirow{3}{*}{\textbf{Methods}} & \multicolumn{4}{c}{\textbf{Java\&Python}} & \multicolumn{4}{c}{\textbf{C\&DS}} & \multicolumn{4}{c}{\textbf{CS\&MA}} \\ 
\cmidrule(lr){2-5} \cmidrule(lr){6-9} \cmidrule(lr){10-13} 
& \multicolumn{2}{c}{Course Java} & \multicolumn{2}{c}{Course Python} & \multicolumn{2}{c}{Course C} & \multicolumn{2}{c}{Course DS} &  \multicolumn{2}{c}{Course CS} & \multicolumn{2}{c}{Course MA} \\
\cmidrule(lr){2-3} \cmidrule(lr){4-5} \cmidrule(lr){6-7} \cmidrule(lr){8-9} \cmidrule(lr){10-11} \cmidrule(lr){12-13}
& ACC & AUC & ACC & AUC & ACC & AUC & ACC & AUC & ACC & AUC & ACC & AUC \\ \midrule
DKT & 0.7988 & 0.7064 & 0.7917 & 0.7128 & 0.7617 & 0.6662 & 0.7906 & 0.6706 & 0.7571 & 0.7002 & 0.8000 & 0.8527 \\  
DKT+ & 0.8085 & 0.7070 & 0.7939 & 0.7198 & 0.7595 & 0.6674 & 0.7933 & 0.6774 & 0.7562 & 0.6971 & 0.7916 & 0.8500 \\ 
IEKT & 0.7983 & 0.7132 & 0.7870 & 0.7250 & 0.7567 & 0.6821 & 0.7924 & 0.6871  & 0.7648 & 0.7013 & 0.8077 & 0.8484 \\ \midrule

Deep\_IRT & 0.7964 & 0.6929 & 0.7833 & 0.7058 & 0.7568 & 0.6626 & 0.7872 & 0.6667  & 0.7440 & 0.6643 & 0.8071 & 0.8457 \\ 
DKVMN & 0.7988 & 0.6955 & 0.7834 & 0.7077 & 0.7541 & 0.6644 & 0.7828 & 0.6694  &0.7433 & 0.6769 & 0.8047 & 0.8438 \\ \midrule    

GIKT & 0.8005 & 0.7088 & 0.7880 & 0.7165 & 0.7535 & 0.6708 & 0.7819 & 0.6800  &0.7456 & \textbf{0.7033} & 0.8014 & 0.8415 \\ \midrule


AKT & \underline{0.8110} & 0.7258 & \underline{0.8035} & \underline{0.7301} & 0.7612 & 0.6783 & 0.7970 & 0.6853 & \underline{0.7679} & 0.6972 & 0.8086 & 0.8447\\ 
simpleKT & 0.7942 & 0.7250 & 0.7933 & 0.7270 & 0.7514 & 0.6825 & 0.7873 & 0.6903  &0.7664 & 0.6992 & 0.8061 &  0.8523\\ 
sparseKT & 0.8098 & 0.7222 & 0.7976 & 0.7262 & \underline{0.7656} & \underline{0.6872} & \underline{0.7981} & 0.6924 &0.7512 & 0.6941 & \underline{0.8103} & \underline{0.8551} \\ 
stableKT & 0.7912 & 0.7194 & 0.7960 & 0.7292 & 0.7577 & 0.6856 & 0.7951 & 0.6850  &0.7505 & 0.6899 & 0.7998 & 0.8504 \\ 

CL4KT & 0.8068 & \underline{0.7260} & 0.7965 & 0.7286 &0.7630 & 0.6768 & 0.7836 & \underline{0.6931} & 0.7584 & 0.6957 & 0.8035 & 0.8506\\ 
DTransformer & 0.8040 & 0.7152 & 0.7947 & 0.7216 & 0.7601 & 0.6680 & 0.7745 & 0.6863 & 0.7510 & 0.6851 & 0.7998 & 0.8469 \\ \midrule


TransKT(Ours) & \textbf{0.8341}* & \textbf{0.7703}* & \textbf{0.8119}* & \textbf{0.7440}* & \textbf{0.7793}* & \textbf{0.7022}* & \textbf{0.8214}* & \textbf{0.7532}*  & \textbf{0.7765}* & \underline{0.7020} & \textbf{0.8222}* & \textbf{0.8629}*\\ \bottomrule
\end{tabular}
\end{adjustbox}

\caption{Performance comparison of TransKT and 12 KT models on three datasets. The best results are in bold, and the second-best results are underlined. $*$ indicates statistical significance over the best baseline, measured by t-test with p-value $ \leq $ 0.01.}
\label{tab:experiment}

\end{table*}

\begin{table*}[t]
\centering

\vspace{-0.5em}

\begin{adjustbox}{max width=0.85\textwidth}
\begin{tabular}{lcccccccccccc}
\toprule
\multirow{2}{*}{\textbf{Methods}} & \multicolumn{2}{c}{\textbf{Java\&Python}} & \multicolumn{2}{c}{\textbf{C\&DS}} & \multicolumn{2}{c}{\textbf{CS\&MA}} \\ 
\cmidrule(lr){2-3} \cmidrule(lr){4-5} \cmidrule(lr){6-7} 
& \multicolumn{1}{c}{Course Java} & \multicolumn{1}{c}{Course Python} & \multicolumn{1}{c}{Course C} & \multicolumn{1}{c}{Course DS} &  \multicolumn{1}{c}{Course CS} & \multicolumn{1}{c}{Course MA} \\
\midrule

AKT*  & 0.7331(+0.73\%)  & 0.7373(+0.72\%)  & 0.6845(+0.62\%)  & 0.6960(+1.07\%)  & 0.7016(+0.44\%)  & 0.8510(+0.63\%)\\ 
stableKT* & 0.7315(+1.21\%)  & 0.7398(+1.06\%)  & 0.6944(+0.88\%) 
& 0.6945(+0.95\%)  & 0.7031(+1.32\%)  & 0.8613(+1.09\%)  \\ 
CL4KT*  & 0.7344(+0.84\%)  & 0.7361(+0.75\%)  & 0.6872(+1.04\%)  & 0.7015(+0.84\%)  & 0.7008(+0.51\%) & 0.8570(+0.64\%)\\ 
\bottomrule
\end{tabular}
\end{adjustbox}
\caption{AUC performance of variant versions of the baseline method across three datasets. * denotes the replacement of each model's original question representation with the one extracted by the proposed semantic-enhanced knowledge transfer module. The values in parentheses indicate the improvement over the original version.}
\vspace{-1em}
\label{tab:semantic}
\end{table*}
We present the details of our experiment settings and the corresponding results in this section. We conduct comprehensive analyses and investigations to illustrate the effectiveness of proposed TransKT model.


\subsection{ Experimental Setup}
\paragraph{Datasets.} We further process the publicly available PTADisc dataset~\cite{hu2023ptadisc} to obtain three sub-datasets specifically tailored to support the analysis of the CCKT task. These are \textit{Java} and \textit{Python} (Java\&Python); \textit{C programming} and \textit{Data Structure and Algorithm Analysis} (C\&DS); \textit{C programming} and \textit{Discrete Mathematics} (CS\&MA). The statistics for the three CCKT datasets are in Table~\ref{tab:dataset_info}. 
\paragraph{Baselines.} We compare TransKT with the state-of-the-art methods, including 1) \emph{deep sequential methods}: DKT~\cite{piech2015deep}, DKT+~\cite{yeung2018addressing}, IEKT~\cite{long2021tracing}; 2) \emph{deep memory-aware methods}: DKVMN~\cite{zhang2017dynamic} and Deep\_IRT~\cite{yeung2019deep}; 3) \emph{graph based method}: GIKT~\cite{yang2021gikt}; 4) \emph{attention based methods}: 
AKT~\cite{ghosh2020context}, 
simpleKT~\cite{liu2023simplekt}, sparseKT~\cite{huang2023towards}, stableKT~\cite{li2024enhancing}; 5) \emph{contrastive learning based methods}: CL4KT~\cite{lee2022contrastive} and DTransformer~\cite{yin2023tracing}. To ensure a fair comparison, we perform joint training using the merged cross-course learning records $\mathcal{R}^{X \cup Y}$ for these baselines.

\paragraph{Experimental Settings and Metrics.} 
We use the AdamW optimizer to train all models, fixing the embedding size at 256 and the dropout rate at 0.3. The learning rate and $L_2$ coefficient are chosen from the sets \{1e-3, 1e-4, 1e-5\} and \{1e-4, 5e-5, 1e-5\}, respectively. The hyperparameters $\eta$ and $\lambda$ are chosen from the range 0.1 to 0.9 with a step size of 0.1. To ensure a fair comparison, method-specific hyperparameters (\textit{e.g.}, Cognition Space Size for IEKT) were set according to the specifications outlined in their respective papers, while we optimized shared hyperparameters, such as the learning rate and $L_2$ regularization coefficient, for all baseline methods. In line with existing KT studies, our evaluation metric includes both AUC and Accuracy (ACC). We repeat each experiment 5 times and report the averaged metrics. In addition, we set an epoch limit of 200 and employ an early stopping strategy if the AUC shows no improvement for 10 consecutive epochs. 

\subsection{Results }
\subsubsection{Overall Performance} 
Table~\ref{tab:experiment} shows the performance comparison of our model with other KT models on three CCKT datasets: Java\&Python, C\&DS, CS\&MA. The results reveal several key observations:
(1) Our TransKT consistently outperforms the baseline models across all datasets, irrespective of the level of course similarity. Whether the datasets exhibit high course similarity (\textit{e.g.}, Java\&Python and C\&DS) or significant differences in course topics (CS\&MA), TransKT demonstrates superior performance. Specifically, it achieves an average increase of 1.51\% and 3.17\% over the best baseline model in ACC and AUC, respectively. This highlights the effectiveness and generalizability of the TransKT.
(2) The performance gains of TransKT vary from course to course and are influenced by the distribution of interleaved interaction records. In datasets like C\&DS, learners tend to engage more in Course C before Course DS, making the knowledge transfer from Course C to Course DS more beneficial. Similarly, in datasets like CS\&MA, Course CS has denser interaction records compared to Course MA, leading to a similar situation.
(3) Among all baseline methods, ranked from weakest to strongest, are deep memory-aware methods, deep sequential methods, graph based method, contrastive learning based methods and attention based methods. AKT performs the best, possibly due to its monotonic attention mechanism modeling forgetting behavior in cross-course learning.

\subsubsection{The Impact of Semantic-Enhanced Knowledge Transfer}
\begin{figure}[h]
  \centering
  \includegraphics[width=\linewidth]{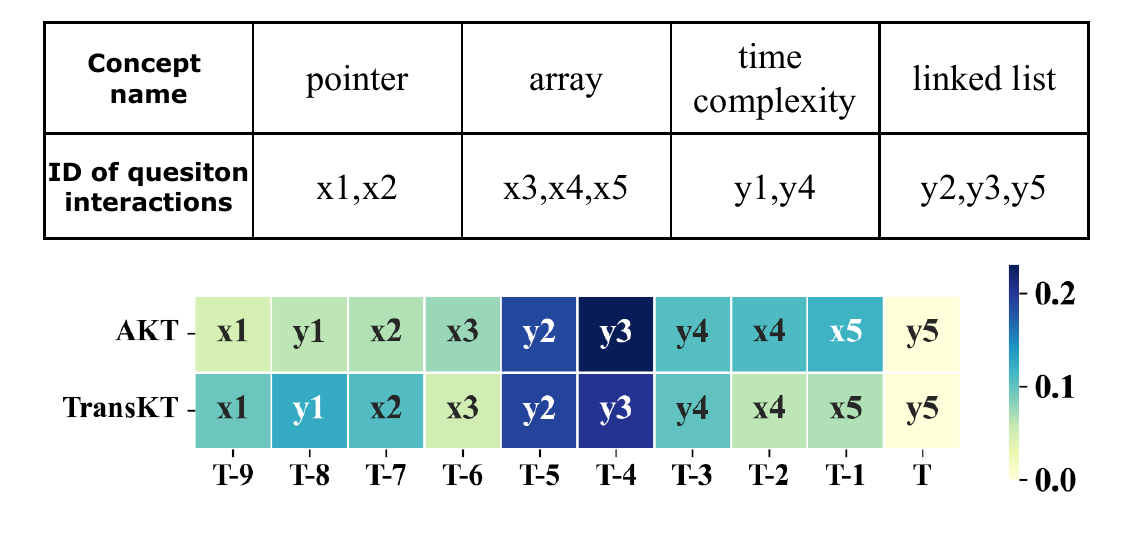}
  \vspace{-2em}
  \caption{A representative example illustrating the benefits of semantic-enhanced knowledge transfer.}
  \vspace{-1em}
  \label{fig:case}
\end{figure}

To further validate the significance of incorporating question and concept semantic features in cross-course scenarios, we select three baselines and enhance them by integrating the cross-course concept graph construction (§\ref{sec:3-1}) and semantic-enhanced knowledge transfer (§\ref{sec:3-2}). As shown in Table~\ref{tab:semantic}, the results indicate that these baselines perform better on the CCKT dataset than their original versions. Specifically, they achieve average improvements of 0.96\% in AUC. This indicates that the design in TransKT for facilitating cross-course knowledge transfer can be seamlessly integrated with other methods, resulting in improved performance on CCKT tasks.
Figure~\ref{fig:case} visualizes the interaction level attention weights of a single learner at time T (\textit{i.e.}, during interaction $y5$) towards interactions from $T-9$ to $T-1$. The AKT model predominantly focuses on interactions $y2$ and $y3$, which share the same concept name (``linked list''), while exhibiting temporal monotonicity in relation to the other interactions. In contrast, TransKT not only emphasizes interactions $y2$ and $y3$ but also extends its focus to interactions $y1$ and $y4$ (``time complexity'') within the same course, as well as cross-course interactions $x1$ and $x2$ (``pointer''). This indicates that TransKT effectively links the learner's current interaction with their previous ones based on the semantic similarity of concept names, thereby enhancing knowledge transfer between interactions. These observations underscore the efficacy of the semantic-enhanced knowledge transfer module.
\subsubsection{The Impact of Cross-Course Contrastive Learning}
\begin{figure}[t]   
    \begin{minipage}[t]{0.49\linewidth}
        \centering
        \includegraphics[width=\textwidth]{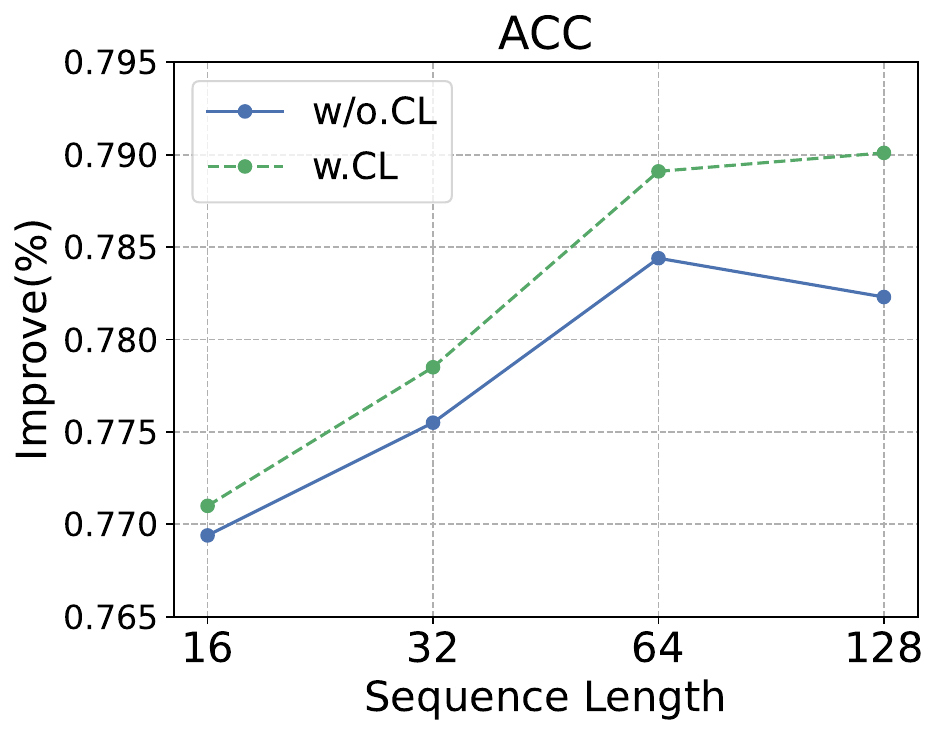}
    \end{minipage}
    \begin{minipage}[t]{0.49\linewidth}
        \centering
        \includegraphics[width=\textwidth]{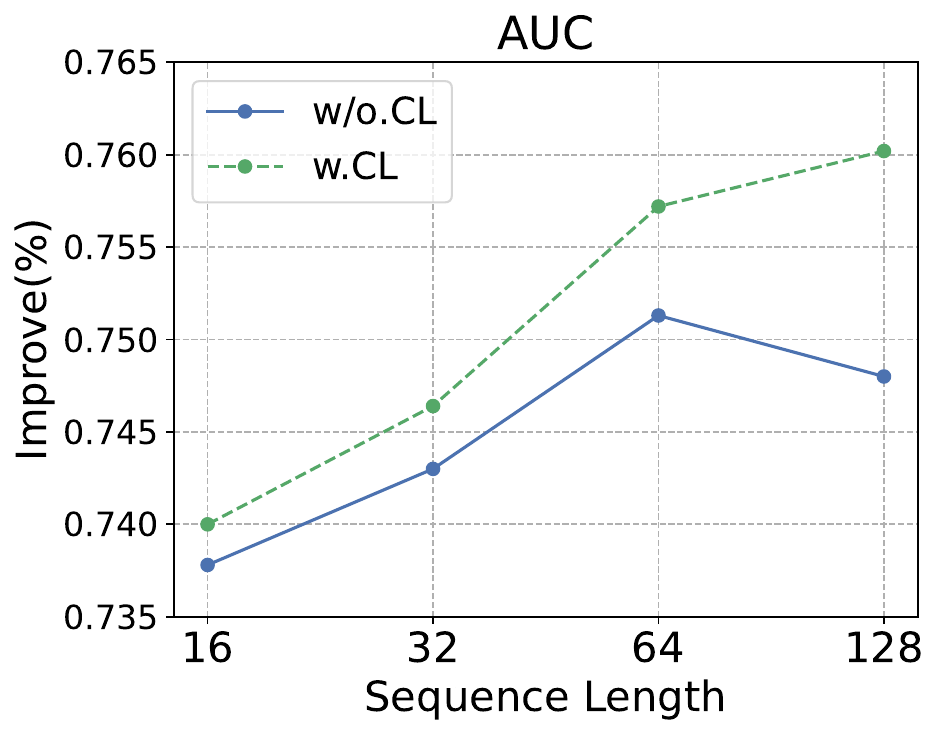}
    \end{minipage}
\vspace{-0.5em}
        \caption{ Performance comparison of cross-course contrastive learning objective across different learning sequence lengths.}
    \label{fig:contrastive}
\end{figure}
The proposed cross-course contrastive learning objective is based on learning history, implying that the performance improvement is correlated with the length of the learning history. To investigate this, we conducted experiments using the Java\&Python dataset, further analyzing the scenarios in which this objective applies. As shown in Table~\ref{tab:dataset_info}, the average learning history length in the Java\&Python dataset is 145.4. In our experiments, we set four different upper limits for the learning history length: [16, 32, 64, 128] (i.e., truncating learner interaction records within these ranges). As depicted in Figure~\ref{fig:contrastive}, the benefits of the cross-course contrastive learning objective increased with the upper limit of learning history length, reaching a 1.22\% improvement in AUC when the upper limit was set to 128. This highlights the importance of aligning learners' knowledge states in single-course and cross-course in long-term learning scenarios.

\subsubsection{Ablation Study} We conduct a comprehensive ablation study on three datasets. We first define the following variations to investigate the impact of each component in TransKT: 1) w/o.KP removes the semantic knowledge propagation module. 2) w/o.SE removes the semantic feature encoding from TransKT and use ID-based embeddings for questions and concepts. 3) w/o.LLM directly extracts semantic features from questions and concepts without using LLM for explanation generation. 4) w/o.CL removes the cross-course contrastive objective from TransKT.
\begin{figure}[t]
  \centering
  \includegraphics[width=\linewidth]{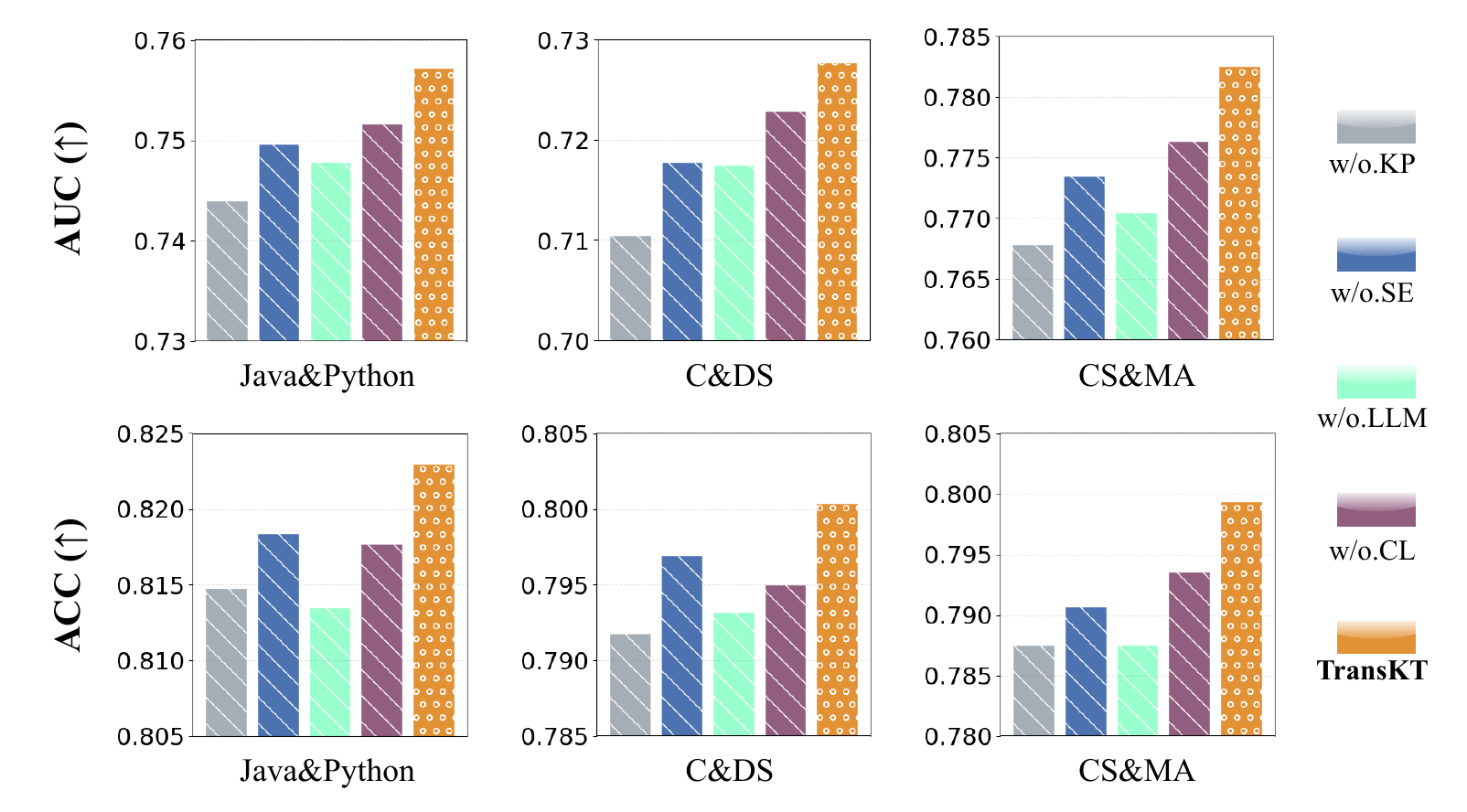}
  \vspace{-2em}
  \caption{Ablation study.}
  \vspace{-1em}
  \label{fig:ablation}
\end{figure}
 The results in Figure~\ref{fig:ablation} reveal several observations: (1) All variants suffer relative performance declines across the three datasets on different metrics, demonstrating the contribution of the designed modules in TransKT. Among them, w/o.KP performs the worst, indicating that establishing a cross-course concept graph for knowledge transfer can effectively uncover the correlation between learners' behaviors across different courses. This is fundamental for effective CCKT. (2) The performance of w/o.LLM is even worse than that of w/o.SE, indicating the challenge of directly extracting semantic features without the interpretative capabilities of LLM. In fact, extracting semantic features directly from the original content of questions and concepts performs even worse than using randomly initialized embedding features. (3) By introducing a cross-course contrastive learning objective, TransKT effectively integrates the learning of both single-course and cross-course knowledge states. This approach not only effectively leverages learning information from multiple courses but also helps to mitigate noise from other courses, enhancing the overall performance.

\section{Conclusion}
In this paper, we introduce TransKT, a novel model for cross-course knowledge tracing (CCKT). TransKT utilizes zero-shot large language model (LLM) queries to construct a comprehensive cross-course concept graph and employs an LLM-to-LM pipeline to enhance semantic features, significantly improving the performance of Graph Convolutional Networks (GCNs) in knowledge transfer. By aligning single-course and cross-course knowledge states through a cross-course contrastive objective, TransKT offers a more robust and comprehensive understanding of learners' knowledge states. Extensive experiments on three real-world datasets demonstrate that TransKT surpasses state-of-the-art KT models in predicting learners' performance across courses.

\section*{Acknowledgements}
This research was partially supported by grants from the National Natural Science Foundation of China (No.62307032, No.62037001), and the ``Pioneer'' and ``Leading Goose'' R\&D Program of Zhejiang under Grant No. 2025C02022.


\bibliographystyle{named}
\bibliography{ijcai25}
\begin{appendices}
\newpage
\newpage

\section{Appendix}
\subsection{Prompt Template}
\begin{figure}[h]
  \centering
  \includegraphics[width=\linewidth]{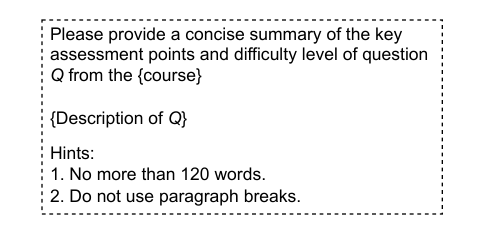}
  \vspace{-2.5em}
  \caption{Prompt template, $pro_{q}$.}
  \vspace{-2em}
  \label{fig:pro_q}
\end{figure}

\begin{figure}[h]
  \centering
  \includegraphics[width=\linewidth]{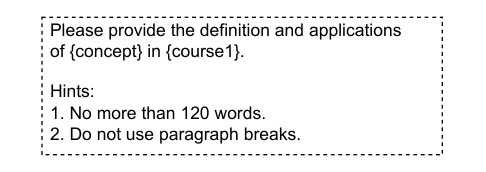}
  \vspace{-2em}
  \caption{Prompt template, $pro_{c}$.}
   \vspace{-2.5em}
  \label{fig:pro_c}
\end{figure}

We selected five questions (and concepts) from the CCKT dataset, generated prompts using the prompt templates shown in Figure~\ref{fig:pro_q} and Figure~\ref{fig:pro_c}, and manually crafted well-structured answers for these prompts to serve as few-shot examples. These examples were then used for few-shot prompting, offering more consistent and high-quality guidance to ensure stable semantic enhancement using the LLM.

\begin{figure}[h]
  \centering
  \includegraphics[width=\linewidth]{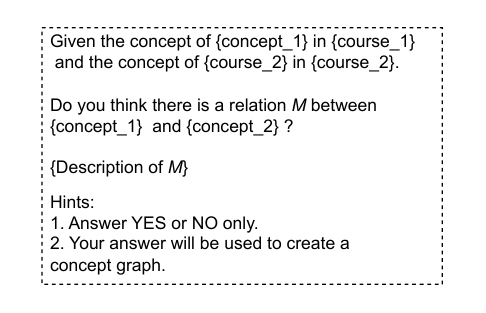}
  \vspace{-2.5em}
  \caption{Prompt template, $pro_{r}$.}
   \vspace{-1em}
  
  \label{fig:pro_r}
\end{figure}

We employ prompt template $pro_{r}$ (Shown in Figure~\ref{fig:pro_r}) for zero-shot link prediction, conducting five independent queries for each of the four candidate relationships. To ensure reliability, we use a majority vote to determine the presence of each relationship.

\subsection{Cross-Course Knowledge Tracing Dataset}
we utilize the publicly available PTADisc
dataset~\cite{hu2023ptadisc} to synthesize three cross-course knowledge tracing datasets. The preprocessing steps are as follows:
(1) Initially, we select learners from the PTADisc dataset who have interaction logs in two courses.
(2) Next, we remove questions from the dataset that have been answered fewer than 10 times to maintain data quality.
(3) Lastly, we exclude learners who have fewer than three interaction records in a single course or fewer than 10 interaction records in a cross-course setting to focus on highly engaged learners.

\subsection{Hybrid Hard Negative Sampling}
The key distinction between proposed hybrid hard negative sampling and previous negative sampling methods lies in its interaction replace strategy. In our constructed dataset, we observe a notable trend: learners who are able to correctly answer questions involving advanced concepts generally also perform well on questions related to basic concepts. However, the reverse does not always hold true. For instance, a learner who can solve questions labeled with ``B+ tree'' may also answer simpler questions like ``linked list'' correctly.Inspired by this, we construct an interaction replace strategy for negative sampling. We assume that the difficulty of a problem is directly reflected by the probability of it being answered correctly in the training dataset. The entire process of the proposed hybrid hard negative sampling is shown in Algorithm~\ref{algo:1}.
To demonstrate the effectiveness of the proposed hybrid hard negative sampling, we conducted further ablation study on Java\&Python dataset. The results are presented in Table~\ref{apd:ablation}, where \textbf{wo.RF+IR} indicates the removal of both hard negative sampling strategies (Response Flip, Interaction Replace).

\begin{table}[h!]
\centering
\begin{adjustbox}{max width=0.92\columnwidth}
\begin{tabular}{lcccc}
\toprule
Method & ACC(X) & AUC(X) & ACC(Y) & AUC(Y) \\
\midrule
wo.RF+IR & 0.8289 ± 0.0013 & 0.7628 ± 0.0017 & 0.8056 ± 0.0005 & 0.7398 ± 0.0006 \\
w.RF     & 0.8278 ± 0.0009 & 0.7679 ± 0.0017 & 0.8089 ± 0.0005 & 0.7401 ± 0.0004 \\
w.IR     & 0.8296 ± 0.0014 & 0.7666 ± 0.0008 & 0.8079 ± 0.0010 & 0.7387 ± 0.0006 \\
w.RF+IR  & 0.8341 ± 0.0006 & 0.7703 ± 0.0014 & 0.8119 ± 0.0008 & 0.7440 ± 0.0011 \\
\bottomrule
\end{tabular}
\end{adjustbox}
\caption{Ablation results of the hybrid hard negative sampling on the Java\&Python dataset (confidence intervals measured by t-test with p-value $ \leq $ 0.01).}
\label{apd:ablation}
\end{table}

\begin{algorithm}[h]
    \caption{Hybrid hard negative sampling}
    \label{algo:1}
    \KwIn{interaction logs $\mathcal{R}$, thresholds $0<\theta_1<\theta_2<1$}
    \KwOut {hard negative samples $\tilde{\mathcal{R}}$}
    
    \BlankLine
    
    $\tilde{\mathcal{R}}=[]$\;
    \For{$(q,r)$ \textbf{in} $\mathcal{R}$}{
        $rand$ =random(0,1)\;
        \If{ $rand<\theta_1$}{
            $\tilde{\mathcal{R}}$.append$(q,1-r)$\    \%response flip strategy\;
        }
        \ElseIf{$\theta_1<rand<\theta_2$}{
            \%interaction replace strategy\;
            \uIf{$r$ = 1}{
                $\tilde{q}=q^{\downarrow}$ \%$q^{\downarrow}$ means a easier question than $q$\;
            }
            \Else{
                $\tilde{q}=q^{\uparrow}$ $\%q^{\uparrow}$ means a harder question than $q$\; 
            }
            $\tilde{\mathcal{R}}$.append$(\tilde{q},1-r)$\;
        }
       \Else{
            $\tilde{\mathcal{R}}$.append$(q,r)$\;
       }
    }
    
    \Return $\tilde{\mathcal{R}}$\;
\end{algorithm}

\begin{figure*}[t]   
    \begin{minipage}[t]{0.24\linewidth}
        \centering
        \includegraphics[width=\textwidth]{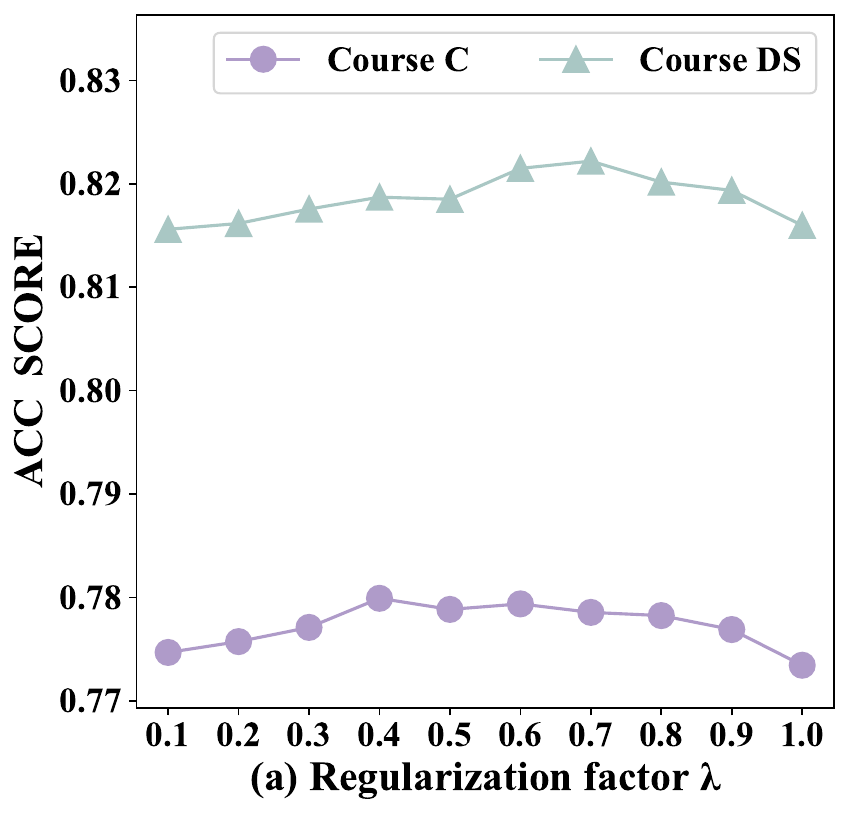}
    \end{minipage}
    \begin{minipage}[t]{0.24\linewidth}
        \centering
        \includegraphics[width=\textwidth]{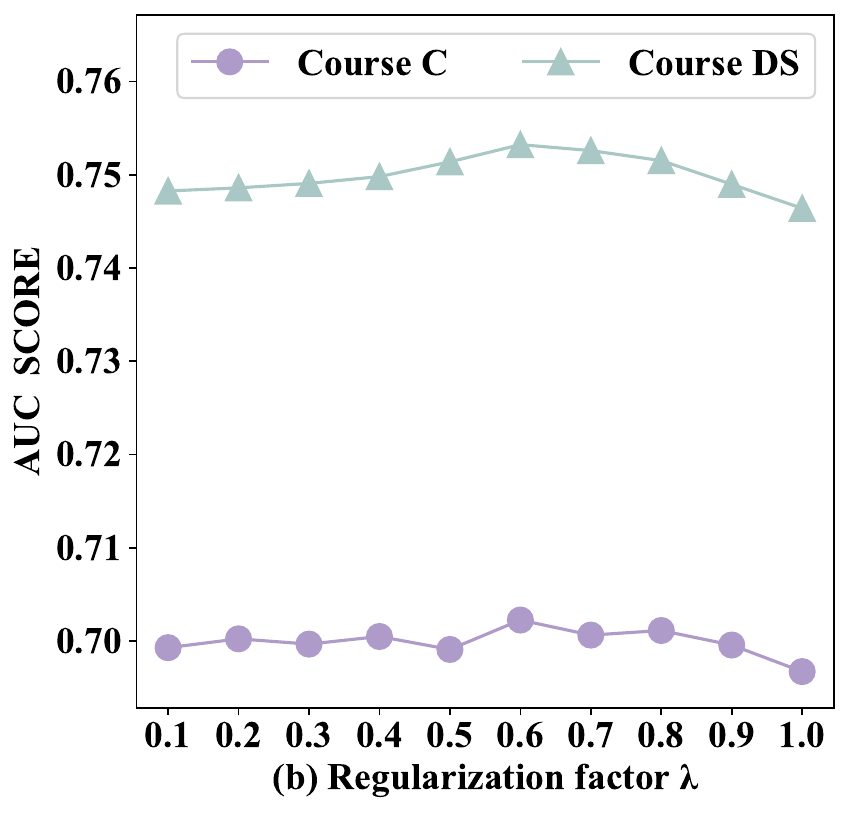}
    \end{minipage}
    \begin{minipage}[t]{0.24\linewidth}
        \centering
        \includegraphics[width=\textwidth]{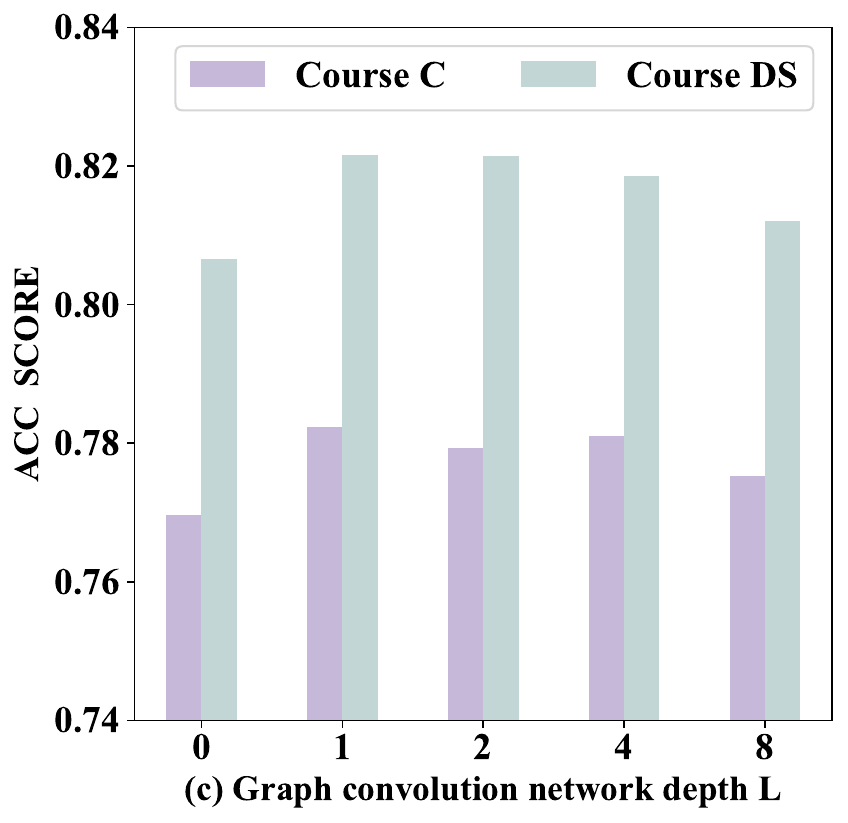}
    \end{minipage}
    \begin{minipage}[t]{0.24\linewidth}
        \centering
        \includegraphics[width=\textwidth]{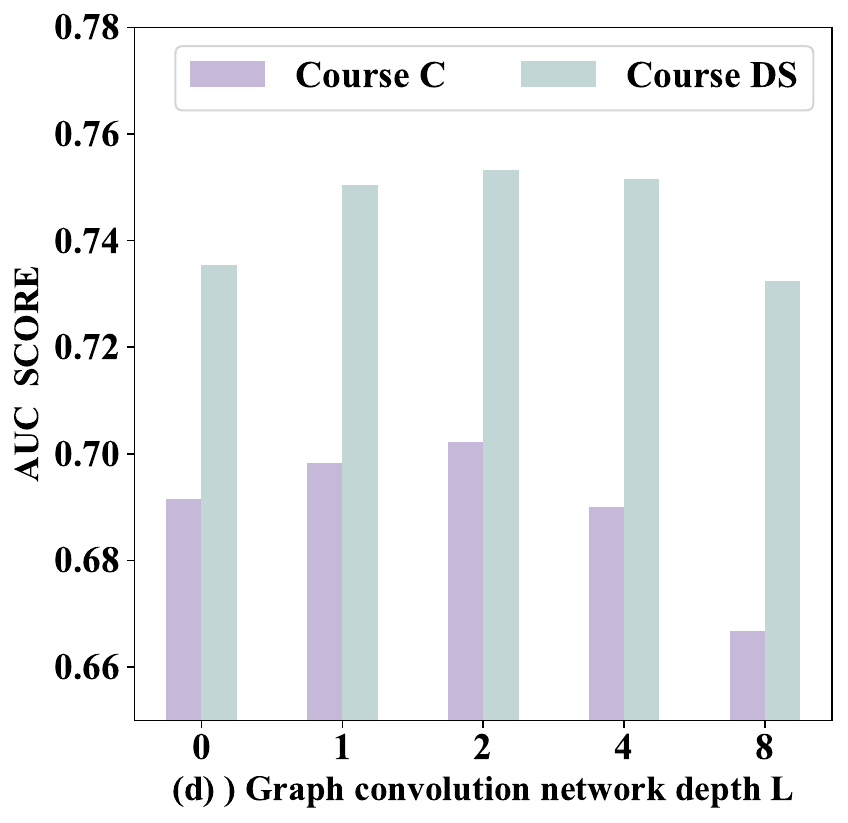}
    \end{minipage}
    \vspace{-0.2cm}
    \caption{ The impact of regularization factor $\lambda$ and graph convolution network depth $L$ on C\&DS dataset.}
    \label{fig:lambda}
    \vspace{-0.2cm}
\end{figure*}

\subsection{Hyperparameter Analysis}
In this section, we conduct experiments based on the C\&DS dataset to explore the sensitivity of TransKT to the regularization factor $\lambda$ graph convolution network depth $L$.
\subsubsection{Impact of regularization factor} For the hyperparameter $\lambda$, Figure.~\ref{fig:lambda}(a) and~\ref{fig:lambda}(b) show the ACC and AUC of TransKT in the C and DS courses, respectively. Our TransKT model demonstrates varying levels of improvement with $\lambda$ values between 0.4 and 0.8, compared to configurations without the cross-course contrastive objective ($\lambda$=1.0). Moreover, a higher $\lambda$ setting implies greater weighting of the predicted loss, which can accelerate model training convergence. Therefore, $\lambda$=0.7 strikes a reasonable balance between model performance and efficiency.
\subsubsection{Impact of graph convolution network depth} For the hyperparameter $L$, Figure.~\ref{fig:lambda}(c) and~\ref{fig:lambda}(d) show the ACC and AUC of TransKT in the C and DS courses, respectively. We report results for $L$=\{0, 1, 2, 4, 8\}, where $L$=0 indicates no semantic-enhanced knowledge propagation module. In general, our model demonstrates stable improvements for $L$=\{1, 2\}, while showing degraded performance for $L$=\{4, 8\}. We attribute this decrease in model performance to over-smoothing caused by increasing the depth of graph convolutional networks. Therefore, we recommend selecting a smaller number of layers, specifically $L$=\{1, 2\}.

\subsection{Implementation Details of Baseline Methods}
\begin{itemize}
     \item DKT~\cite{piech2015deep} is the first deep KT model that uses an LSTM layer to encode the learners’ knowledge state for predicting their responses.
    \item DKT+~\cite{yeung2018addressing} introduces regularization terms corresponding to reconstruction and waviness in the loss function of the original DKT model to enhance the consistency of KT predictions.
    \item DKVMN~\cite{zhang2017dynamic} employs a key-value memory network, utilizing static and dynamic matrices to respectively store relationships between latent concepts and update learners' knowledge states based on question interactions.
    \item Deep\_IRT~\cite{yeung2019deep} is a combination of item response theory (IRT)~\cite{de2004explanatory} and DKVMN, and is an explainable knowledge tracing model based on deep learning.
    \item AKT~\cite{ghosh2020context} leverages an attention mechanism to model the temporal relationship between questions and learners' past interactions, utilizing a Rasch~\cite{rasch1993probabilistic} model to regularize concept and question representation, thus enhancing knowledge tracing by discriminating questions on the same concept.
    \item GIKT~\cite{yang2021gikt} innovatively employs a graph convolutional network to integrate question-skill correlations and model the dynamic interactions among learners' exercise histories, current state, target questions, and related skills, improving knowledge tracing accuracy.
    \item IEKT~\cite{long2021tracing} is a model that estimates learners' understanding of questions before predicting their responses and assesses learners' sensitivity to knowledge acquisition for a given question before updating their knowledge state.
    \item simpleKT~\cite{liu2023simplekt} tackles the knowledge tracing task by incorporating question-specific variations through the Rasch model. It employs a standard dot-product attention function to extract time-aware information from interactions.
    \item sparseKT~\cite{huang2023towards} enhances classical scaled dot-product attention by incorporating a k-selection module to selectively pick items with the highest attention scores, extracting influential historical interactions to estimate learners' knowledge state.
    \item CL4KT~\cite{lee2022contrastive} enhances the AKT model by incorporating a contrastive learning framework and four data augmentation techniques, improving knowledge tracing by effectively handling data sparsity and differentiating between similar and dissimilar learning histories.
    \item DTransformer~\cite{yin2023tracing} introduces a novel architecture and training paradigm in KT by employing Temporal and Cumulative Attention mechanisms and a contrastive learning algorithm, enhancing the accuracy and stability of tracing learners' evolving knowledge states from question-level mastery.
    \item stableKT~\cite{li2024enhancing} is able to learn from short sequences, maintain stable and consistent performance when generalizing on long sequences, and capture hierarchical relationships between questions and their associated concepts.    
\end{itemize}
To ensure a fair comparison between these baselines and TransKT in a cross-course setting, we conducted joint training using the merged cross-course learning records $\mathcal{R}^{X \cup Y}$ and individually recorded the predictive performance for each course.

\end{appendices}

\end{document}